\documentclass[twocolumn,a4paper]{article}
\usepackage[colorlinks, urlcolor=blue,
	pdffitwindow, bookmarksopen,
	filecolor=blue,
	bookmarksopenlevel=0,
	pdfpagelayout=SinglePage, driverfallback=dvipdfm]{hyperref}
\usepackage{ncsp23} 
\usepackage{color}
\usepackage{cite}
\usepackage{amsmath,amsfonts}
\usepackage{algorithmic}
\usepackage{multirow}
\usepackage{latexsym}
\usepackage{newtxtext,newtxmath}
\usepackage{algorithm}
\usepackage{enumitem}
\usepackage{here}
\usepackage{cite}
\usepackage{graphicx}
\usepackage{textcomp}
\usepackage{xcolor}
\usepackage{subcaption}
\usepackage{caption}
\begin{document}

\title{Combined Use of Federated Learning and Image Encryption \\ for Privacy-Preserving Image Classification with Vision Transformer}

\name{
 \begin{tabular}{c}
Teru Nagamori$^1$ and Hitoshi Kiya$^1$
 \end{tabular}
 }

\address{
\begin{tabular}{c}
$^1$Tokyo Metropolitan University\\
6-6 Asahigaoka, Hino-shi, Tokyo 191-0065, Japan\\
E-mail: nagamori-teru@ed.tmu.ac.jp, kiya@tmu.ac.jp
\end{tabular}
}

\maketitle

\section*{Abstract}
In recent years, privacy-preserving methods for deep learning have become an urgent problem. Accordingly, we propose the combined use of federated learning (FL) and encrypted images for privacy-preserving image classification under the use of the vision transformer (ViT). The proposed method allows us not only to train models over multiple participants without directly sharing their raw data but to also protect the privacy of test (query) images for the first time. In addition, it can also maintain the same accuracy as normally trained models. In an experiment, the proposed method was demonstrated to well work without any performance degradation on the CIFAR-10 and CIFAR-100 datasets.

\vspace{-8px}
\section{Introduction}
It has been very popular for data owners to train and test deep neural network (DNN) models in cloud environments. However, data privacy such as personal medical records may be compromised in cloud environments, so privacy-preserving methods for deep learning have become an urgent problem \cite{kiya2022overview, E, F}.\par
One of the solutions is to use federated learning (FL) \cite{FL, federated-learning}, which was proposed by Google. FL is capable of significantly preserving clients’ private data from being exposed to adversaries.
However, FL aims to construct models over multiple participants without directly sharing their raw data, so the privacy of test (query) images is not considered.\par
Another approach is to encrypt a trained model and then encrypt test (query) images are applied to the encrypted models \cite{KIYA20232022MUI0001, maung_privacy, jimaging8090233, D}. However, this approach does not consider constructing models over multiple participants without directly sharing their raw data, although the visual information of test images can be protected.\par
For these reasons, we propose a method for the combined use of FL and encrypted test images for privacy-preserving image classification with the vision transformer (ViT) \cite{ViT}. The proposed method allows us not only to train models over multiple participants without directly sharing their raw data but to also protect the privacy of test (query) images for the first time. In addition, it can maintain the same accuracy as that of models normally trained with plain images.

\vspace{-8px}
\section{Related Work}
\subsection{Federated Learning (FL)}
Federated Learning (FL) \cite{FL, federated-learning} is the scheme proposed by Google, in which multiple data owners can collaborate on training statistical models. In FL, multiple developers send only parameters of models, which they have trained in their local environments, to a server. On the server, the sent parameters are integrated, and the integrated parameters are sent back to the developers. This process is repeated to train models. Therefore, there is no need to directly share their raw data to concentrate them on a server, so FL allows us to carry out privacy-preserving model training with distributed resources \cite{9069945, 9777682}. However, FL has not considered the protection of test data so far.\par
Two types of model integration methods for FL were proposed in \cite{federated-learning}. One is the FederatedSGD algorithm (FedSGD), which computes the average of the gradients over the inputs to the model. The other is the FederatedAveraging algorithm (FedAVG), which aggregates the trained weight parameters of each client and averages them.

\begin{figure}[tb]
    \centering
    \includegraphics[bb=0 0 1123 1417,scale=0.21]{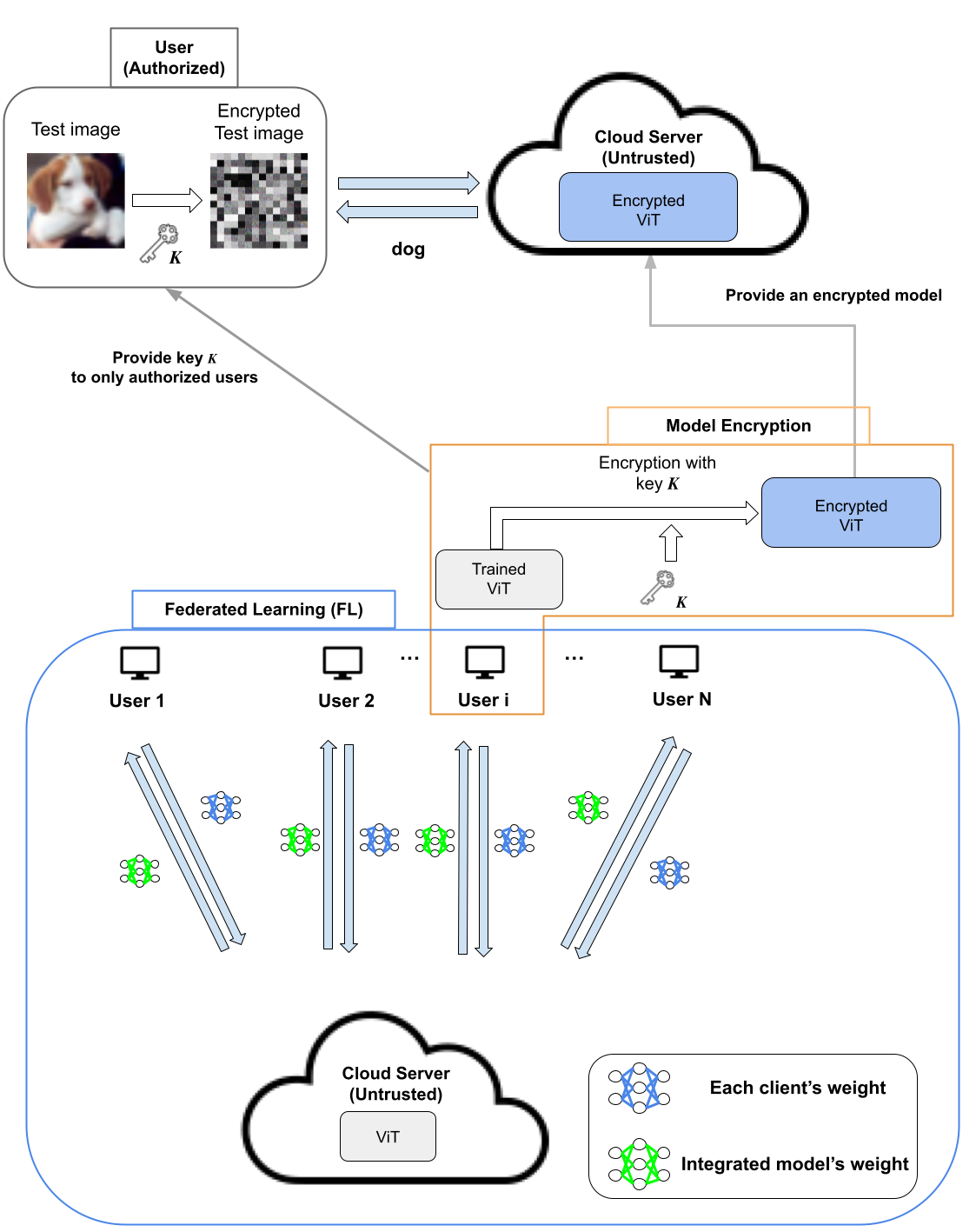}
    \caption{Overview of proposed method}
    \label{Overview}
\end{figure}

\subsection{Privacy-preserving deep learning}
Many researchers have studied privacy-preserving deep learning to protect visual information on plain images. A typical solution is to use learnable encryption that allows us to apply encrypted images for testing and training models. Privacy-preserving learning methods with learnable encryption are classified into two classes: methods for protecting the privacy of test images \cite{jimaging8090233}, and methods for protecting the privacy of both test and training images \cite{maung_privacy}. In particular, several methods for protecting the privacy of test images can maintain the same accuracy as that of plain images when using an isotropic network such as ViT \cite{jimaging8090233}. However, these conventional methods with learnable encryption have not considered training models over multiple participants.

\subsection{Vision Transformer (ViT)}
The Vision Transformer (ViT) \cite{ViT} is generally used for image classification tasks, and it is known for its high classification performance. In ViT, a transformer encoder is used instead of a CNN where an image is divided into patches and every patch is transformed into a one-dimensional learnable vector for processing. The architecture of ViT has two embeddings: position embedding to maintain information about where the cropped patches are located in an image, and patch embedding to transform each patch into a learnable vector. \par
In this paper, we focus on these two embedding structures so that the privacy of test images is protected without any performance degradation.

\begin{figure}[tb]
    \centering
    \begin{minipage}{4truecm}
      \centering
      \includegraphics[bb=0 0 224 224,scale=0.4]{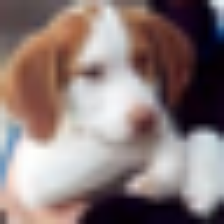}
      \subcaption{Original \\ ($224 \times 224 \times 3$)}
    \end{minipage}
    \begin{minipage}{4truecm}
      \centering
      \includegraphics[bb=0 0 224 224,scale=0.4]{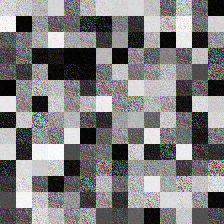}
      \subcaption{Encrypted image \\ (block size = 16)}
    \end{minipage}

    \caption{An example of encrypted images with proposed method}
    \label{encrypted_image}
\end{figure}

\vspace{-8px}
\section{Proposed Method}

\subsection{Overview}
As shown in Fig. \ref{Overview}, in the proposed method, a ViT model is trained by using FL over multiple participants without directly sharing their raw data. Each user can then encrypt the trained model with an independent secret key that each user manages by his/herself to protect the privacy of test (query) images. \par
In ViT, an input image $x \in \mathbb{R}^{h \times w \times c}$ is segmented into $N$ patches with a size of $p \times p$, where $h$, $w$ and $c$ are the height, width, and number of channels of the image.
In addition, $N$ is given as $hw/p^2$. After that, each patch is flattened as $x_{p}^i = [x_{p}^i(1), x_{p}^i(2),\dots , x_{p}^i(L)]$. Finally, the embedding patch is given as
\begin{align}
    z_{0} =& [x_{class}; x_{p}^1\mathbf{E}; x_{p}^2\mathbf{E}; \dots x_{p}^i\mathbf{E}; \dots x_{p}^N\mathbf{E}] + \mathbf{E_{pos}},
\end{align}
where 
\begin{align*}
\mathbf{E_{pos}} =& ((e_{pos}^0)^\top (e_{pos}^1)^\top \dots (e_{pos}^i)^\top \dots (e_{pos}^N)^\top)^\top,\\
  L =& p^{2}c, x_{class} \in \mathbb{R}^{D}, \ x_{p}^{i} \in \mathbb{R}^{L}, \ e_{pos}^i \in \mathbb{R}^{D},\\
  \mathbf{E} \in& \mathbb{R}^{L \times D}, \ \mathbf{E_{pos}} \in \mathbb{R}^{(N+1) \times D},
\end{align*}
$x_{class}$ is the classification token, $\mathbf{E}$ is the embedding (patch embedding) to linearly map each patch to dimensions $D$, $\mathbf{E_{pos}}$ is the embedding (position embedding) that gives position information to patches in the image, $e_{pos}^0$ is the information of the classification token, $e_{pos}^i$ is the position information of each patch, and $z_0$ is an embedded patch.\par
In this paper, we propose not only to encrypt test images but to encrypt the two embeddings: patch embedding $\mathbf{E}$ and position embedding $\mathbf{E_{pos}}$, in a trained model.

\subsection{Model Encryption}
In the proposed method, patch embedding $\mathbf{E}$ and position embedding $\mathbf{E_{pos}}$ of ViT are encrypted by random matrices generated by secret keys, respectively. 

\subsubsection{Patch Embedding Encryption}
In the proposed method, the following transformation matrix $\mathbf{E_{a}}$ is used for patch embedding encryption.\par
    \begin{align}
        \mathbf{E_{a}} =
        \begin{bmatrix} 
              k_{(1,1)} & k_{(1,2)} & \dots & k_{(1,L)} \\
              k_{(2,1)} & k_{(2,2)} &\dots  & k_{(2,L)} \\
              \vdots & \vdots & \ddots & \vdots \\
              k_{(L,1)} & k_{(L,2)} & \dots & k_{(L,L)}\\
        \end{bmatrix},
    \end{align}
    \begin{align*}
       \mathbf{E_{a}} &\in \mathbb{R}^{L \times L}, \ \mathrm{det} \ \mathbf{E_{a}} \neq 0 \\
       \ k_{(i,j)} &\in \mathbb{R},\ i,j \in \left\{1, \dots, L\right\}.
    \end{align*}

Note that the element values of $\mathbf{E_{a}}$ are randomly decided but $\mathbf{E_{a}}$ has to have an inverse matrix. \par
Then, by multiplying $\mathbf{E}$ by $\mathbf{E_{a}}$, an encrypted patch embedding $\mathbf{\hat{E}}$ is given by 
\begin{equation}
    \mathbf{\hat{E}} =\mathbf{E_{a}}\mathbf{E}.
\end{equation}

\subsubsection{Position Embedding Encryption}
A position embedding encryption method is carried out as below.
\begin{itemize}
    \item[1)]Generate a random integer vector with a length of $N$ as 
        \begin{equation}
            l_{t} = [l_{e}(1), l_{e}(2), \dots, l_{e}(i)\ , \dots ,l_{e}(N)] \ ,
        \end{equation}
        where
        \begin{align*}
            l{e}(i) \in& \left\{1,2,...,N\right\}, \\
            le(i) \neq& le(j) \ \text{if} \ i \neq j \\
            i,j \in& \left\{1, \dots, N\right\}.
        \end{align*}
    \item[2)] Given $m_{(i,j)}$ as
        \begin{equation}
            m_{(i,j)} = \left\{ \begin{matrix}0&(j\neq l_{e}(i))\\ 1&(j = l_{e}(i)) \end{matrix} \right. .
        \end{equation}
    \item[3)] Define a random matrix as
    \begin{align}
        \mathbf{E_{b}} =
        \begin{bmatrix} 
              1 & 0 & 0 & \dots & 0 \\
              0 & m_{(1,1)} & m_{(1,2)} & \dots & m_{(1,N)} \\
              0 & m_{(2,1)} & m_{(2,2)} &\dots  & m_{(2,N)} \\
              \vdots & \vdots & \vdots & \ddots & \vdots \\
              0 & m_{(N,1)} & m_{(N,2)} & \dots & m_{(N,N)}\\
        \end{bmatrix},
    \end{align}
    \begin{align*}
       \mathbf{E_{b}} &\in \mathbb{R}^{(N+1) \times (N+1)}.
    \end{align*}
    For instance, if $N$ = 3 and $l_{t}$ = [1,3,2], $\mathbf{E_{b}}$ is given by 
        \begin{align}
        \mathbf{E_{b}} =
            \begin{bmatrix} 
                  1 & 0 & 0 & 0 \\
                  0 & 1 & 0 & 0 \\
                  0 & 0 & 0 & 1 \\
                  0 & 0 & 1 & 0 \\          
            \end{bmatrix}.
        \end{align}
    \item[4)] Transform $\mathbf{E_{pos}}$ to $\mathbf{\hat{E}_{pos}}$  as
        \begin{equation}
            \mathbf{\hat{E}_{pos}} =\mathbf{E_{b}}\mathbf{E_{pos}}.
        \end{equation}
    \end{itemize}

\subsection{Test Image Encryption}
The procedure for encrypting an test image is shown below.
\begin{itemize}

    \item[(a)] Divide an image $x \in \mathbb{R}^{h \times w \times c}$ into blocks with a size of $p \times p$ such that $B =  \left\{B_{1},\dots ,B_{N}\right\}$.
    
    \item[(b)] Generate permutated blocks $\hat{B}$ by
    \begin{equation}
                \hat{B} = B\mathbf{E_{b}}, \ \hat{B} \in \mathbb{R}^{1 \times N},
    \end{equation}
    For instance, it is expressed as
    \begin{equation}
               \left\{\hat{B}_{1}, \hat{B}_{2}, \dots ,\hat{B}_{N}\right\} = \left\{B_{3}, B_{N}, \dots ,B_{1}\right\},
    \end{equation}
    
    \item[(c)]Flatten each block $\hat{B_{i}}$ into a vector $b_{i}$ such that
     \begin{equation}
        b_{i} = [b_{i}(1), \dots , b_{i}(L)].
    \end{equation}

    where $b_{i}$ is equal to $x_{p}^i$ in Eq. (1).
    
    \item[(d)]Calculate an encrypted vector $\hat{b}_{i}$ by
    \begin{equation}
                \hat{b}_{i} = b_{i}\mathbf{E}_{a}^{-1}, \ \hat{b}_{i} \in \mathbb{R}^{1 \times L}.
    \end{equation}
    
    It is also expressed as
    \begin{equation}
                \hat{b}_{i} = b_{i}\mathbf{E}_{a}^{-1} = x_{p}^i\mathbf{E}_{a}^{-1} = \hat{x}_{p}^i.
    \end{equation}
    
    \item[(e)]Concatenate the encrypted vectors $\hat{b}_{i} (i \in {1, \dots, N})$ into an encrypted test image $\hat{x}$.

\end{itemize}

When using an encrypted test image and encrypted embeddings, Eq. (1) can be expressed as follows.
\begin{align}
    z_{0} =& [x_{class}; \hat{x}_{p}^3\mathbf{E_{a}}\mathbf{E}; \hat{x}_{p}^N\mathbf{E_{a}}\mathbf{E}; \dots \hat{x}_{p}^1\mathbf{E_{a}}\mathbf{E}] + \mathbf{\hat{E}_{pos}} \notag \\
    =& \mathbf{E_{b}}[x_{class}; x_{p}^1\mathbf{E}; x_{p}^2\mathbf{E}; \dots x_{p}^N\mathbf{E}] + \mathbf{\hat{E}_{pos}}
\end{align}\par
From Eqs. (8) and (14), the proposed method can avoid the influence of encryption.\par
Figure \ref{encrypted_image} shows an example of images encrypted with this procedure,  where $k(i,j)$ in Eq. (2) was determined in the same way as the procedure for generating $m_{(i,j)}$ in Eq. (6), and $p$ (patch (block) size) was 16.

\vspace{-8px}
\section{Experimental Results}
\subsection{Setup}
Experiments were conducted on the CIFAR-10 and CIFAR-100 datasets, where images were resized from $32 \times 32 \times 3$ to $224 \times 224 \times 3$ because we used ViT pre-trained with ImageNet-1K as a model.
For training models with FL, 10 clients were assumed where each client had 5,000 training images and 1,000 test images. Also, we used FedAVG \cite{federated-learning} as the method of model integration.\par
Models were trained using the stochastic gradient descent (SGD) with an initial learning rate of $10^{-3}$, a momentum of 0.9, and a batch size of 8. We also used the cross-entropy loss function. In addition, models were integrated every epoch, and the total number of epochs was set to 10. After the tenth integration, the integrated model was encrypted with secret keys, and every client used the secret keys to encrypt their test images. \par

\begin{table}[tbh]
 \caption{Classification accuracy of proposed method}
 \label{table:cifar10}
 \centering
 \scalebox{0.8}{
  \begin{tabular}{|c|cc|}
   \hline
    & Integrated Model & Baseline \\
    \hline
   CIFAR-10 & 0.977 & 0.978\\
   \hline
   CIFAR-100 & 0.851 & 0.851\\
   \hline
  \end{tabular}
    }
\end{table}

\subsection{Image Classification Performance}
We evaluated the performance of models in terms of classification accuracy. Table \ref{table:cifar10} shows experimental results on the CIFAR-10 and CIFAR-100 datasets, which have 10 and 100 classes, respectively. ”Integrated Model” indicates results when encrypted test images were applied to encrypted integrated models, and ”Baseline” represents results when plain test images were applied to plain models normally trained with plain images.\par
From the results, the combined use of FL and encrypted images was verified to give the same accuracy as that of models normally trained with plain images. Accordingly, the proposed method allows us not only to train models over multiple participants without directly sharing raw data but to also protect the visual information of test images.

\vspace{-8px}
\section{Conclusions}
In this paper, we proposed the combined use of FL learning and encrypted images for the first time. In the experiments, the proposed method was demonstrated to well work without any performance degradation on the CIFAR-10 and CIFAR-100 datasets.

\vspace{-8px}
\section*{Acknowledgment}
This study was partially supported by JSPS KAKENHI (Grant Number JP21H01327).
\bibliographystyle{IEEEtran.bst}
\bibliography{ref.bib}

\begin{thebibliography}{10}
\providecommand{\url}[1]{#1}
\csname url@samestyle\endcsname
\providecommand{\newblock}{\relax}
\providecommand{\bibinfo}[2]{#2}
\providecommand{\BIBentrySTDinterwordspacing}{\spaceskip=0pt\relax}
\providecommand{\BIBentryALTinterwordstretchfactor}{4}
\providecommand{\BIBentryALTinterwordspacing}{\spaceskip=\fontdimen2\font plus
\BIBentryALTinterwordstretchfactor\fontdimen3\font minus
  \fontdimen4\font\relax}
\providecommand{\BIBforeignlanguage}[2]{{%
\expandafter\ifx\csname l@#1\endcsname\relax
\typeout{** WARNING: IEEEtran.bst: No hyphenation pattern has been}%
\typeout{** loaded for the language `#1'. Using the pattern for}%
\typeout{** the default language instead.}%
\else
\language=\csname l@#1\endcsname
\fi
#2}}
\providecommand{\BIBdecl}{\relax}
\BIBdecl

\bibitem{kiya2022overview}
H.~Kiya, M.~AprilPyone, Y.~Kinoshita, S.~Imaizumi, and S.~Shiota, ``An overview
  of compressible and learnable image transformation with secret key and its
  applications,'' \emph{APSIPA Transactions on Signal and Information
  Processing}, vol.~11, no. 1, e11, 2022.

\bibitem{E}
I.~Nakamura, Y.~Tonomura, and H.~Kiya, ``Unitary transform-based template
  protection and its application to l2-norm minimization problems,''
  \emph{IEICE Transactions on Information and Systems}, vol. E99.D, no.~1, pp.
  60--68, 2016.

\bibitem{F}
W.~Sirichotedumrong, T.~Chuman, S.~Imaizumi, and H.~Kiya, ``Grayscale-based
  block scrambling image encryption for social networking services,'' in
  \emph{2018 IEEE International Conference on Multimedia and Expo (ICME)},
  2018, pp. 1--6.

\bibitem{FL}
\BIBentryALTinterwordspacing
J.~Konečný, H.~B. McMahan, F.~X. Yu, P.~Richtárik, A.~T. Suresh, and
  D.~Bacon, ``Federated learning: Strategies for improving communication
  efficiency,'' \emph{arXiv}, 2016. [Online]. Available:
  \url{https://arxiv.org/abs/1610.05492}
\BIBentrySTDinterwordspacing

\bibitem{federated-learning}
B.~McMahan, E.~Moore, D.~Ramage, S.~Hampson, and B.~A.~y. Arcas,
  ``Communication-efficient learning of deep networks from decentralized
  data,'' in \emph{20th International Conference on Artificial Intelligence and
  Statistics (AISTATS)}, vol.~54.\hskip 1em plus 0.5em minus 0.4em\relax Fort
  Lauderdale, Florida,USA. JMLR: W\&CP, 2017, pp. 1273--1282.

\bibitem{KIYA20232022MUI0001}
H.~Kiya, R.~Iijima, M.~AprilPyone, and Y.~Kinoshita, ``Image and model
  transformation with secret key for vision transformer,'' \emph{IEICE
  Transactions on Information and Systems}, vol. E106.D, no.~1, pp. 2--11,
  2023.

\bibitem{maung_privacy}
M.~AprilPyone and H.~Kiya, ``Privacy-preserving image classification using an
  isotropic network,'' \emph{IEEE MultiMedia}, vol.~29, no.~2, pp. 23--33,
  2022.

\bibitem{jimaging8090233}
\BIBentryALTinterwordspacing
H.~Kiya, T.~Nagamori, S.~Imaizumi, and S.~Shiota, ``Privacy-preserving semantic
  segmentation using vision transformer,'' \emph{Journal of Imaging}, vol.~8,
  no.~9, 2022. [Online]. Available:
  \url{https://www.mdpi.com/2313-433X/8/9/233}
\BIBentrySTDinterwordspacing

\bibitem{D}
H.~Ito, Y.~Kinoshita, M.~Aprilpyone, and H.~Kiya, ``Image to perturbation: An
  image transformation network for generating visually protected images for
  privacy-preserving deep neural networks,'' \emph{IEEE Access}, vol.~9, pp.
  64\,629--64\,638, 2021.

\bibitem{ViT}
A.~Dosovitskiy \emph{et~al.}, ``An image is worth 16x16 words: Transformers for
  image recognition at scale,'' in \emph{9th International Conference on
  Learning Representations, {ICLR} 2021, Virtual Event, Austria, May
  3-7}.\hskip 1em plus 0.5em minus 0.4em\relax OpenReview.net, 2021.

\bibitem{9069945}
K.~Wei \emph{et~al.}, ``Federated learning with differential privacy:
  Algorithms and performance analysis,'' \emph{IEEE Transactions on Information
  Forensics and Security}, vol.~15, pp. 3454--3469, 2020.

\bibitem{9777682}
J.~Zhao \emph{et~al.}, ``Pvd-fl: A privacy-preserving and verifiable
  decentralized federated learning framework,'' \emph{IEEE Transactions on
  Information Forensics and Security}, vol.~17, pp. 2059--2073, 2022.

\end{thebibliography}

\end{document}